\documentclass{cta-author}

\usepackage{comment}
\usepackage{multirow}
\usepackage{booktabs}
\usepackage{here}
\usepackage{hhline}
\usepackage[subrefformat=parens]{subcaption}

{}
{}
{}

\begin{document}

\supertitle{Brief Paper}

\title{Influence of Image Classification Accuracy on Saliency Map Estimation}

\author{\au{Taiki Oyama$^1$} \au{Takao Yamanaka$^1$}}
\address{\add{1}{Department of Information \& Communication Sciences, Sophia University, 7-1 Kioi-cho, Chiyoda-ku, Tokyo, 102-0094, Japan}
\email{oyamat4@gmail.com, takao-y@sophia.ac.jp}}

\begin{abstract}
\looseness=-1 Saliency map estimation in computer vision aims to estimate the locations where people gaze in images. Since people tend to look at objects in images, the parameters of the model pretrained on ImageNet for image classification are useful for the saliency map estimation. However, there is no research on the relationship between the image classification accuracy and the performance of the saliency map estimation. In this paper, it is shown that there is a strong correlation between image classification accuracy and saliency map estimation accuracy. We also investigated the effective architecture based on multi scale images and the upsampling layers to refine the saliency-map resolution. Our model achieved the state-of-the-art accuracy on the PASCAL-S, OSIE, and MIT1003 datasets. In the MIT Saliency Benchmark, our model achieved the best performance in some metrics and competitive results in the other metrics.
\end{abstract}

\maketitle

\section{Introduction}\label{sec:intro}
\begin{figure}[t]
\begin{center}
  \includegraphics[scale=0.248]{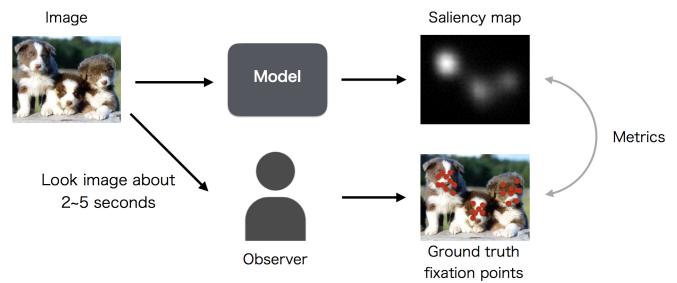}
\end{center}
\caption{Saliency map estimation}
\label{fig:flow}
\end{figure}

Saliency map estimation in computer vision aims to estimate the locations where people gaze in images. The input of the saliency-map estimation task is an image, whereas the output is the saliency map, which is defined as the probability density function of human gaze on the input image, as shown in Figure \ref{fig:flow}. The model to estimate the saliency map is constructed based on the fixations measured with observers for the target images using an eye tracking system. The fixations of several observers are merged to build the set of the ground truth fixation points, and to create the optimal saliency map by blurring. The estimated saliency maps are evaluated with several metrics based on the ground truth fixation points or the optimal saliency maps. The saliency maps are expected to be used for many applications, including image compression \cite{imagecompression}, video compression \cite{videocompression}, image retargeting \cite{imageretargeting}, image cropping \cite{imagecropping}, and virtual/augmented reality \cite{vr}. In addition, modeling the process of human fixations would be useful for understanding the human visual attention \cite{visunderstand}.

In the last few years, convolutional neural networks (CNN) have been widely used for various computer vision tasks such as image classification \cite{alexnet, vgg, resnet, squeezenet, densenet, resnext, efficient-densenet, dpn}, object detection \cite{rcnn, fasterrcnn, yolo}, and semantic segmentation \cite{fcn, dilatedcnn, deeplab}. Many models for estimating saliency maps based on CNN also have been proposed \cite{edn, deepgaze1, saliconnet, deepfix, deepgaze2, dsclrcn, samresnet, densesal}. While it has been difficult for conventional methods \cite{itti, aim, gbvs, sun, dva, sig, aws} based on manually designed low-level features to estimate saliency maps for images containing complex scenes, the models based on CNN can extract effective features from images for the saliency map estimation, leading to the better performance than conventional models. 

It is known that the parameters of the models pretrained on ImageNet are useful for saliency map estimation \cite{deepgaze2, densesal}. This would be because a human tends to look at the centers of objects \cite{objectcenterbias}, which are learned to be recognized in the pretrained model for the ImageNet classification task. However, there is no research on the relationship between the image classification accuracy and performance of saliency map estimation. In this paper, the influence of image classification accuracy on saliency map estimation was investigated. Moreover, the effective architectures to estimate the saliency maps were examined based on multi scale images and upsampling layers to refine the saliency-map resolution. The contributions of this paper are as follows:
\begin{itemize}
\setlength{\leftskip}{0.3cm}
\item The influence of image classification accuracy on saliency map estimation was studied. It was found that there is a strong correlation between image classification accuracy and saliency map estimation accuracy.
\item The effectiveness of upsampling layers and the multipath architecture was investigated. It is shown that there is no need for upsampling layers, while the multipath architecture is useful for saliency map estimation.
\item Our model achieved the state-of-the-art accuracy for saliency map estimation in the PASCAL-S, OSIE, and MIT1003 datasets. In addition, our model achieved the best performance in some metrics for the MIT Saliency Benchmark, and the competitive results in the other metrics.
\end{itemize}
\noindent It is noted that this paper is the extended version of our previous paper in ACPR 2017 \cite{densesal}. Compared with the previous paper, the results on the influence of image classification accuracy on the saliency map estimation were added, in addition to the evaluation results on the MIT Saliency Benchmark. Although the model based on DenseNet \cite{densenet, efficient-densenet} has achieved the state-of-the-art performance in the ACPR 2017 paper, this additional study led to even better performance using the model based on Dual Path Networks (DPN) \cite{dpn}.

This paper is organized as follows. In Section 2, reviews of the saliency map estimation and the image classification are provided. The architecture of the proposed models, the experimental setup, and the experimental results are explained in Section 3, Section 4, and Section 5, respectively. The conclusions are described in Section 6.

\section{Related work}\label{sec:related_work}

\subsection{Conventional methods for saliency map estimation}\label{sec:related_work:conventional_saliencymap}

In the last 20 years, many saliency models have been proposed to estimate the locations in images which attract attentions of humans. Most of the conventional models have used low-level features extracted by edge-detectors and color filters or local image statistics. For example, the model proposed by Itti et al. \cite{itti} extracts early visual features such as intensity channels, color channels, and orientation channels. These visual features are normalized and integrated. AIM \cite{aim} uses coefficients of the basis calculated by ICA in local image patches. The distribution of the coefficients is estimated by the kernel density estimation, which is used for estimating saliency maps based on self-information of the local patches. GBVS \cite{gbvs} exploits channel-wise feature maps computed by linear filtering followed by a nonlinear transformation. To estimate saliency maps, the feature maps are transformed into activation maps and normalized by using the fully-connected directed graph of the feature maps. 

SUN \cite{sun} applies a bayesian framework using local feature maps to estimate saliency maps. The probability distribution over features is learned not from each individual test image but from statistics calculated over the training set of natural images. In DVA \cite{dva}, the incremental coding length (ICL) using features of local image patches is proposed to maximize the entropy of the sampled visual features. In ICL, unexpected features elicit entropy gain in the perception state and are therefore assigned high energy. The probability function of feature activities of this model is updated dynamically. SIG \cite{sig} introduced a simple image descriptor referred to as the image signature. This descriptor can be used to approximate the spatial location of a sparse foreground hidden in a spectrally sparse background. AWS \cite{aws} is based on the whitening of low-level features and has shown good performance for saliency map estimation. 

Although many saliency map models have been proposed using low-level, hand-crafted features as reviewed above, it is difficult to estimate saliency maps of images containing complex scenes \cite{scene} and to select the effective low-level features of images for saliency map estimation.

\subsection{ImageNet classification}\label{sec:related_work:imagenet_classification}

In recent years, many models based on convolutional neural networks (CNN) have been proposed and used for various computer vision tasks such as image classification \cite{alexnet, vgg, resnet, squeezenet, densenet, resnext, efficient-densenet, dpn}, object detection \cite{rcnn, fasterrcnn, yolo}, and semantic segmentation \cite{fcn, dilatedcnn, deeplab}. AlexNet \cite{alexnet} achieved the winning top-5 test error rate of 15.3\% in ILSVRC 2012 competition \cite{imagenet}, compared to 26.2\% achieved by the second-best entry. This model has 8 learned layers --- five convolution layers and three fully connected layers. Each of them except for the last fully connected layer is followed by the activation function of ReLU to prevent the activations from saturating. To reduce overfitting in the fully-connected layers, AlexNet uses the regularization method called dropout and local response normalization. SqueezeNet \cite{squeezenet} has been proposed as smaller CNN architecture. SqueezeNet has several Fire modules: a module is comprised of a squeeze convolution layer which has only 1$\times$1 filters, and an expand layer which has a mix of 1$\times$1 and 3$\times$3 convolution filters. SqueezeNet achieved AlexNet-level accuracy on ImageNet with 50 times fewer parameters. 

VGG \cite{vgg} obtained the first and the second places in the localization and classification tasks in ILSVRC 2014, respectively. The model has small (3$\times$3) convolution filters which enable the model to increase depth (up to 19 weight layers) effectively. ResNet \cite{resnet} won the 1st place on the ILSVRC 2015 classification task. In general, although CNN models with deeper layers have higher performance for various tasks, the problem of vanishing/exploding gradients is likely to occur. In order to avoid it, in ResNet, the residual learning framework using identity mapping by shortcuts is proposed. Their deep residual nets are easy to be optimized, and produced results substantially better than the previous networks by greatly increasing the depth. On the ImageNet dataset, the residual nets were evaluated with the depth of up to 152 layers --- 8 times deeper than VGG but still having lower complexity. 

DenseNet \cite{densenet, efficient-densenet} has shown the competitive results on the four object recognition benchmarks (CIFAR-10, CIFAR-100, SVHN, and ImageNet). In contrast to ResNet, DenseNet combines features by concatenating them along channels. This architecture encourages feature reuse and substantially reduces the number of parameters. Dual Path Networks (DPN) \cite{dpn} achieved the state-of-the-art accuracy on the ImageNet classification task when this model was proposed. ResNet enables feature re-usage while DenseNet enables new features exploration, which are both important for learning good representations. To take the benefits from both path topologies, DPN shares common features while maintaining the flexibility to explore new features through the dual path architectures. Since DPN achieved the high accuracy with the low computational cost and the low GPU memory consumption, this model is useful not only for research but also for real-world applications.

Fully convolutional neural networks (FCN) \cite{fcn} have been proposed for semantic segmentation and have achieved the state-of-the-art accuracy. Different from the architecture for image classification, FCN replaces the fully connected layers to the convolution layers. In recent years, FCN becomes the standard method for pixel level prediction such as semantic segmentation, instance segmentation, and saliency map estimation. 

\subsection{Saliency map estimation with CNN}\label{sec:related_work:cnn_saliency}

CNN models can extract effective features of images for the saliency map estimation automatically. First, eDN \cite{edn} has applied CNN to the task of the saliency map estimation. In 2015, DeepGaze\,I \cite{deepgaze1} has achieved higher performance than eDN by using the weights of AlexNet trained for the ImageNet object recognition. This model has shown that the parameters of the model pretrained on ImageNet are useful for saliency map estimation. This would be because a human tends to look at the centers of objects \cite{objectcenterbias}, which are learned to be recognized in the pretrained model for the ImageNet classification task.

Many models have also applied the transfer learning approaches using deep features since DeepGaze\,I was proposed. In contrast to DeepGaze\,I, SaliconNet \cite{saliconnet} and DeepFix \cite{deepfix} use FCN based on the VGG architecture which has shown better performance than AlexNet on the image classification task. SaliconNet estimates saliency maps with two image scales to allow the model to have robustness to the size of objects in images. DeepFix has dilated convolution filters \cite{dilatedcnn} to enlarge the receptive field and inception modules to capture multi-scale information. DeepGaze\,I\hspace{-.1em}I \cite{deepgaze2} is also based on the VGG architecture, which uses a center-bias layer to incorporate the prior distribution, and log-likelihood as the loss function. Note that the VGG features of DeepGaze\,I\hspace{-.1em}I are not fine-tuned with fixation datasets. These models have improved the performance over DeepGaze\,I in the MIT Saliency Benchmark \cite{mit300}.

DSCLRCN \cite{dsclrcn} and SAM-ResNet \cite{samresnet} use ResNet and achieved high accuracy in the saliency map estimation task. Both models use long short-term memory (LSTM) \cite{lstm} for saliency map estimation in the different ways. In DSCLRCN, LSTM sweeps both horizontally and vertically across the image. This mimics the cortical lateral inhibition mechanisms in the human visual system to incorporate global contexts to assess the saliency of each image location. DSCLRCN uses the features extracted by the model pretrained for scene recognition. In contrast to DSCLRCN, SAM-ResNet uses LSTM to compute an attention map. The feature map extracted by ResNet is input to Attentive Convolutional LSTM that focuses on the most salient regions of the input image to iteratively refine the predicted saliency map. Dilated convolution \cite{deeplab} has been applied to ResNet in both models. DenseNet is used in the model, DenseSal, proposed in our ACPR 2017 paper \cite{densesal} to extract effective feature maps for saliency map estimation. 

In these papers, the performance of the saliency map estimation was evaluated for various architectures based on the CNN models of AlexNet, VGG, ResNet, and DenseNet, though the comparison between the CNN models is scarce. Moreover, it is difficult to compare the CNN models from the published papers because the implementation details in addition to the experimental setups are diversified such as architectures, datasets, objective functions, and optimization methods. Therefore, in this paper, the influence of image classification accuracy on the saliency map estimation is investigated with 15 CNN models including AlexNet, VGG, SqueezeNet, ResNet, DenseNet, and DPN under the same architecture and experimental setups.

\begin{figure*}[t]
  \begin{center}
    \includegraphics[width=18cm, bb=0 0 1120 488]{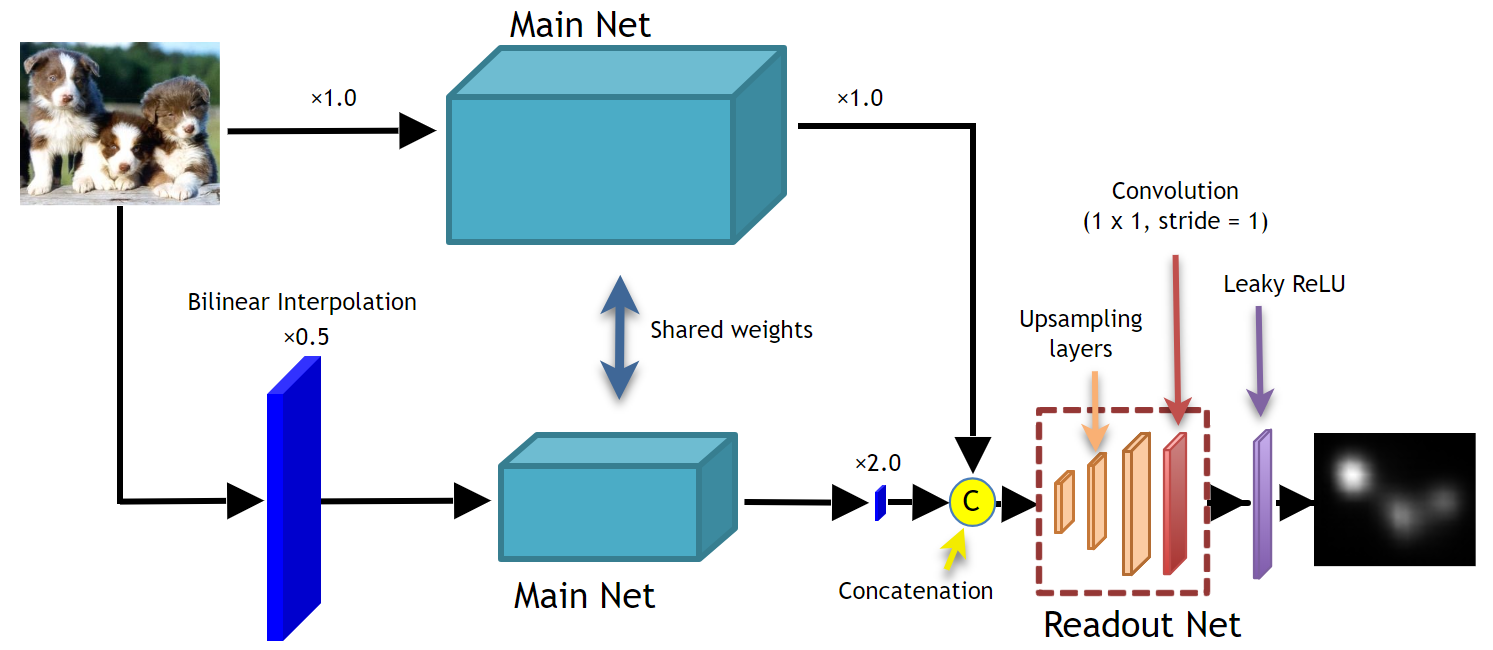}
  \end{center}
  \caption{Architecture of fully convolutional neural networks for saliency map estimation}
  \label{fig:arch}
\end{figure*}

\section{Methods}\label{sec:methods}

\subsection{Architecure}\label{sec:methods:arch}

In this paper, we used the architecture based on SaliconNet \cite{saliconnet}, which has shown high accuracy for estimating saliency maps. The architecture of our model for the saliency map estimation is shown in Figure \ref{fig:arch}. The model consists of two parts, Main Net and Readout Net. In the model, an image is processed at two scales: the original image and the half size of the image obtained by downsampling of bilinear interpolation. These images at two scales are separately processed by fully convolutional neural networks which share weights (Main Nets). This multipath architecture enables the model to be robust to the size of objects in images. The outputs of the two paths are concatenated along channels after resizing by bilinear interpolation. These concatenated feature maps are fed into Readout Net to output the saliency map for the input image.

In the original SaliconNet model \cite{saliconnet}, the fully convolutional neural networks based on VGG-16 \cite{vgg} is used as Main Net, while a $1\times1$ convolution layer as Readout Net. In order to examine the influence of image classification accuracy on saliency map estimation, 15 models were tested as Main Net. The models are based on AlexNet, VGG, ResNet, SqueezeNet, DenseNet, and DPN. The feature maps before the last global pooling layer or the fully connected layers of each CNN model are extracted to be input to Readout Net. 

Several types of upsampling neural networks were examined as Readout Net, because the output size of the fully convolutional neural network is smaller than the original input image. The size of extracted feature maps of the CNN models except for SqueezeNet is about 1/32 while that of SqueezeNet is about 1/16, although the size was increased by adjusting the stride of a pooling layer or a convolution layer in some proposed models. Readout Net examined is a multi-layer neural network composed of bilinear interpolation layers (BI), deconvolution layers (DC) \cite{deconvolution}, or sub-pixel convolution layers (SPC) \cite{spc}, followed by a $1\times1$ convolution layer and Leaky ReLU \cite{leakyrelu}. Each upsampling layer resizes feature maps twice, resulting in saliency maps in the higher resolution. The $1\times1$ convolution layer predicts the saliency map from all the concatenated feature maps from the two Main Nets ($\times$1.0, $\times$0.5). When the number of upsampling layers is $N$, the size of the predicted saliency map by Readout Net is $2^N$ times of the feature maps from Main Net. Note that $N=0$ means Readout Net corresponds to a $1\times1$ convolution layer, same as Readout Net in the original SaliconNet.

\subsection{Components of Main Net}\label{sec:methods:main_net}

\begin{figure*}[t]
\begin{center}
	\begin{tabular}{ccc}
	\begin{minipage}{1\textwidth}
		\centering
		\includegraphics[scale=0.485]{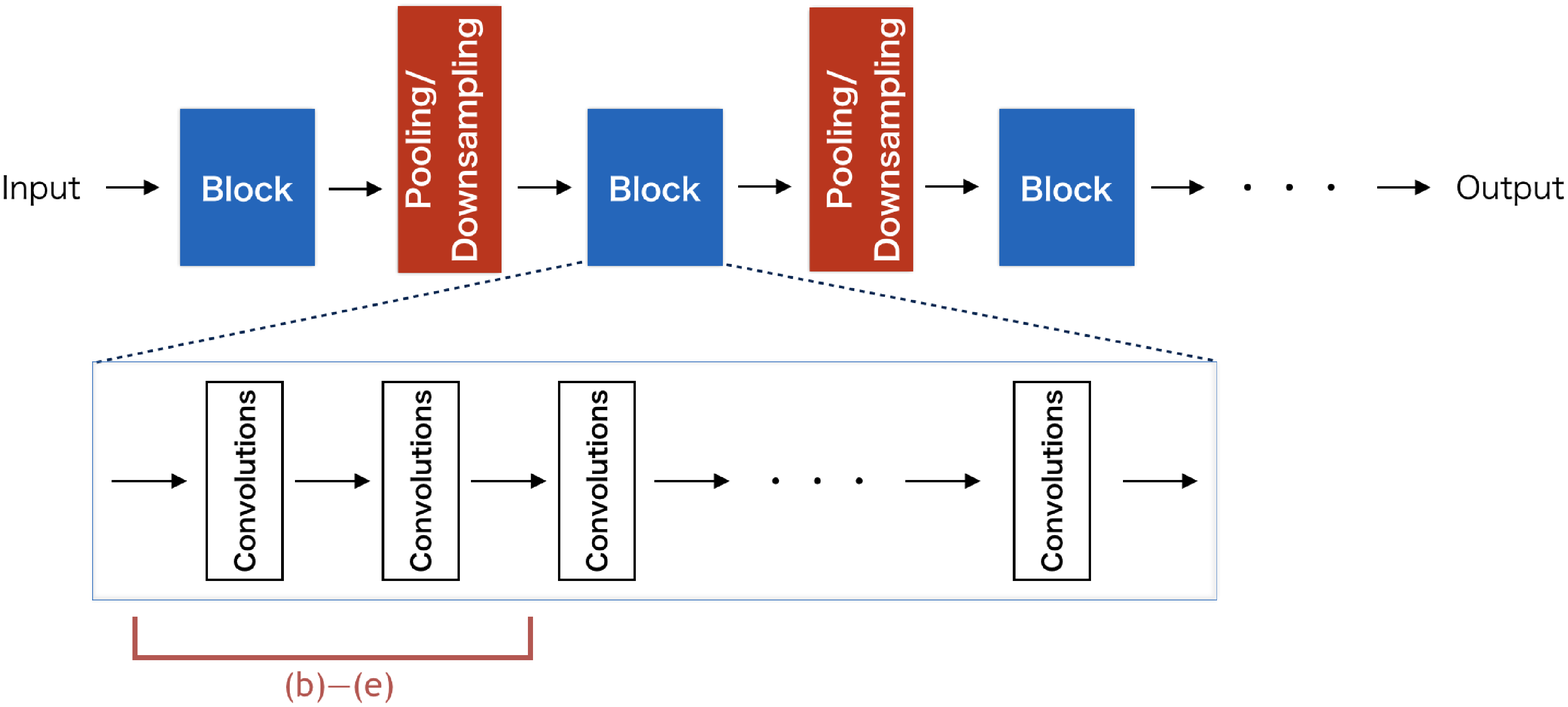}
		\subcaption{Architecture of Main Net}
		\label{}
	\end{minipage}\\
	\begin{minipage}{.5\textwidth}
		\centering
		\includegraphics[scale=0.25]{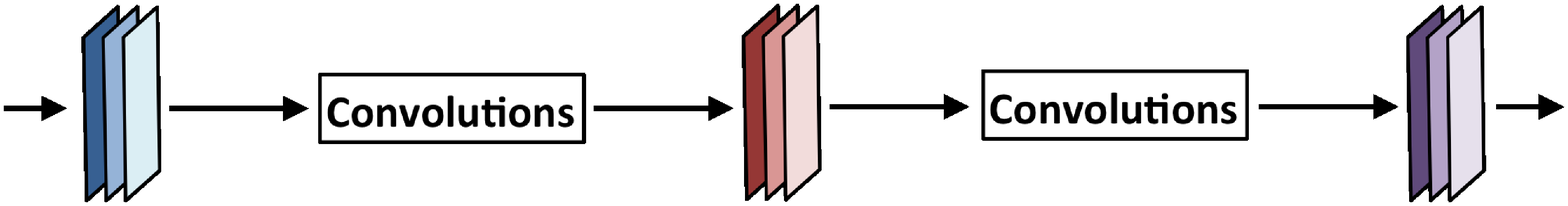}
		\subcaption{Standard CNN such as AlexNet, VGG, and SqueezeNet}
		\label{}
	\end{minipage}
		\begin{minipage}{.5\textwidth}
		\centering
		\includegraphics[scale=0.25]{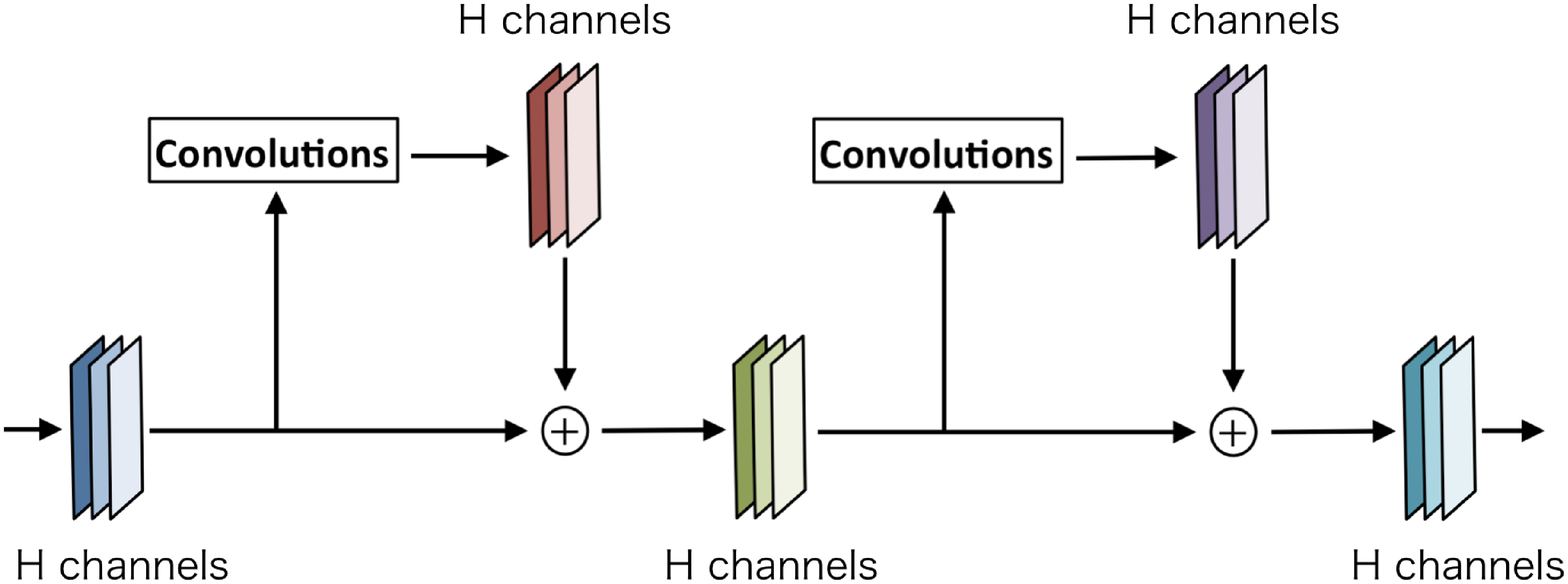}
		\subcaption{ResNet}
		\label{}
	\end{minipage}\\
	\begin{minipage}{.5\textwidth}
		\centering
		\includegraphics[scale=0.25]{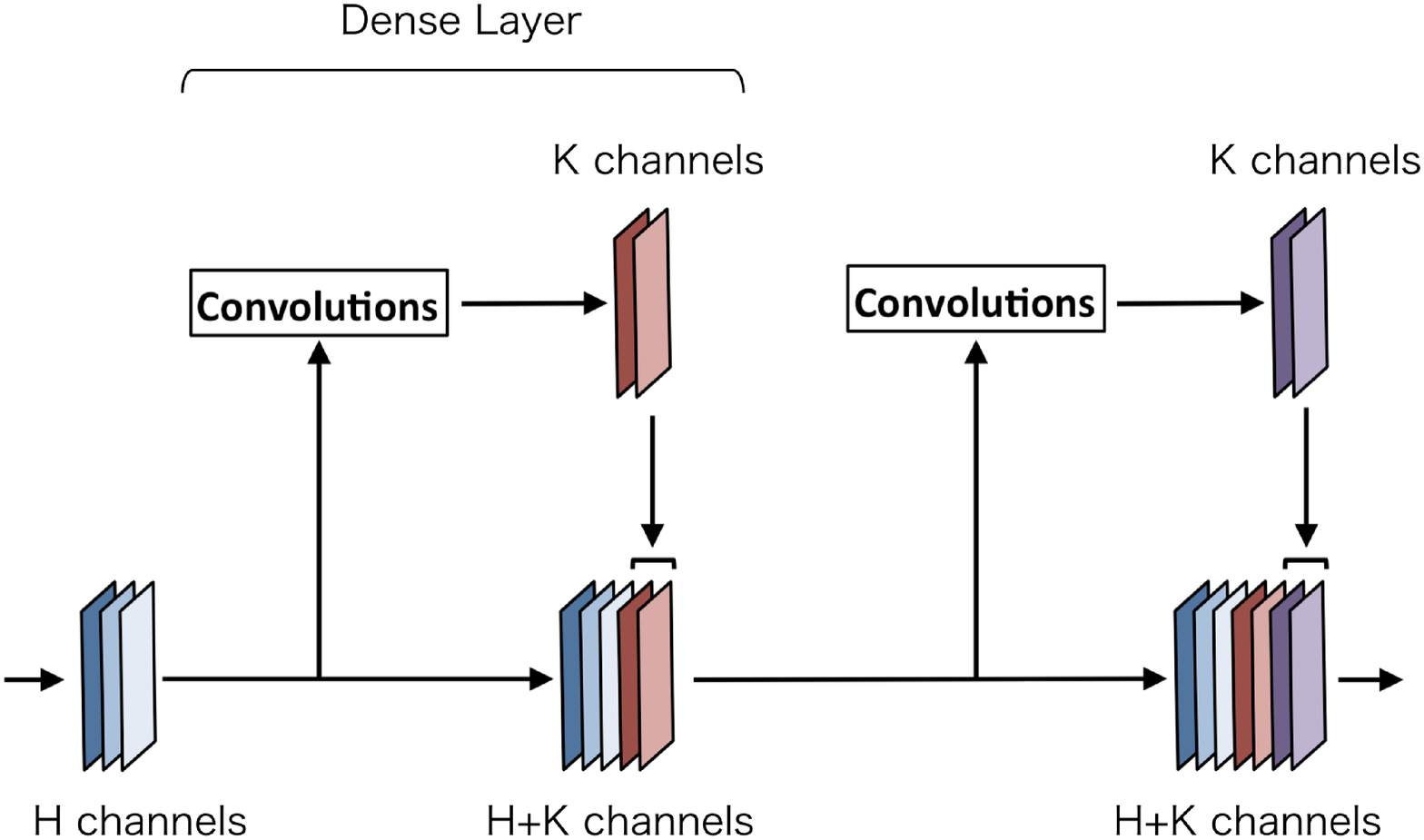}
		\subcaption{DenseNet (K=2)}
		\label{}
	\end{minipage}
		\begin{minipage}{.5\textwidth}
		\centering
		\includegraphics[scale=0.25]{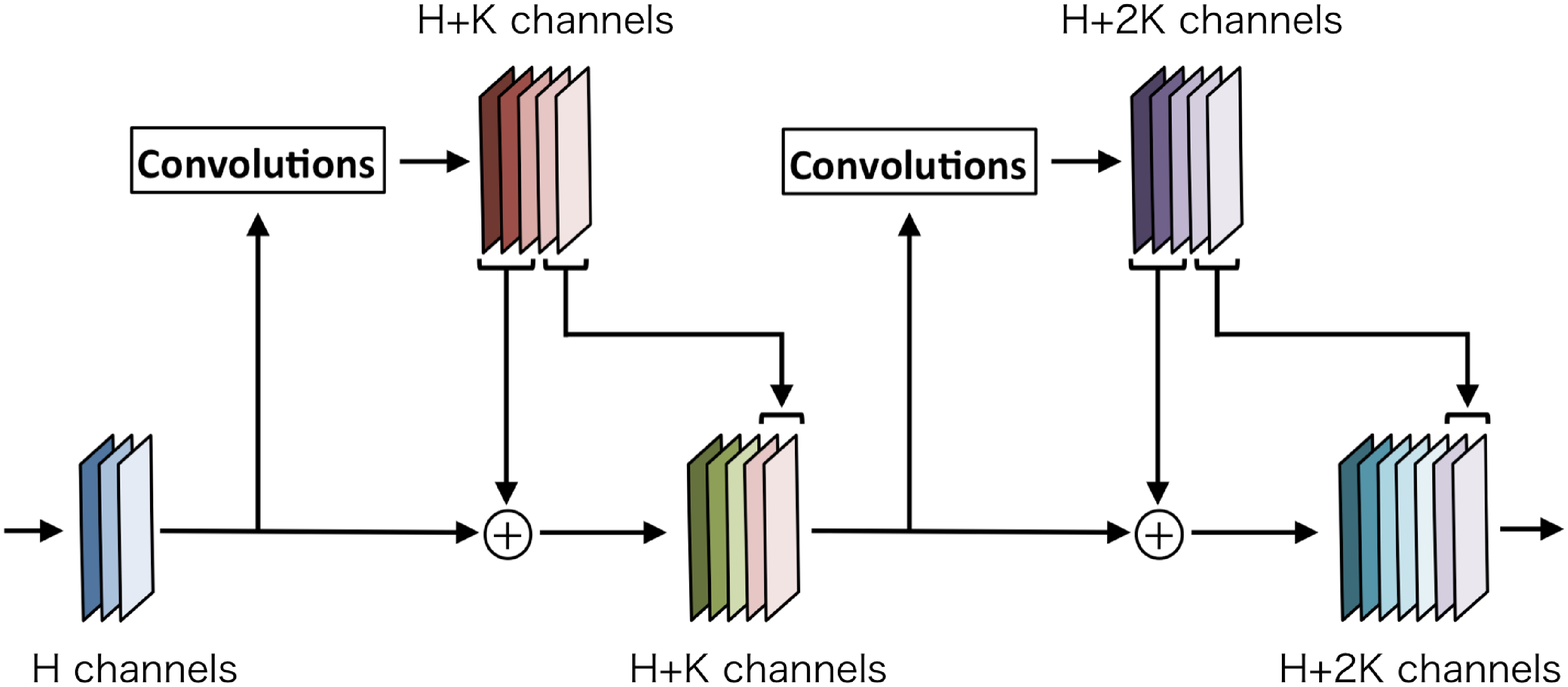}
		\subcaption{DPN (K=2)}
		\label{}
	\end{minipage}
	\end{tabular}
\end{center}
\caption{Comparison of models in Main Net. K: growth rate}
\label{fig:main-archs}
\end{figure*}

The models pretrained on ImageNet for the image classification task were used as Main Net for investigating influence of image classification accuracy on saliency map estimation. The models used in this paper include AlexNet, VGG, ResNet, SqueezeNet, DenseNet, and DPN. The architecture of these models is shown in Figure \ref{fig:main-archs}. Each model has several pooling or donwsampling layers between blocks, as shown in Figure \ref{fig:main-archs} (a). Each block consists of several Convolutions as shown in Figure \ref{fig:main-archs} (a), each of which is composed of convolution layers and activation functions such as ReLU with/without batch normalization layers \cite{bn}. The architectures within the block are different among the models, as shown in Figure \ref{fig:main-archs} (b)-(e).

\begin{itemize}
\setlength{\leftskip}{0.3cm}
\item {\bf AlexNet} \cite{alexnet} has 8 learned layers. These layers are 5 convolution layers and 3 fully connected layers. The output of the fifth layer (the last convolution layer) is extracted to be input to Readout Net. As shown in Figure \ref{fig:main-archs} (b), AlexNet is based on the standard architecture without skip connections such as ResNet. 
\item {\bf VGG} \cite{vgg} has achieved the deep layered neural networks up to 19 layers, by using small ($3\times3$) convolution filters instead of large size convolution filters. By stacking several $3\times3$ convolution layers, the receptive field of the convolution can be enlarged with fewer parameters: for example, the stack of three $3\times3$ convolution layers has the effective receptive field of $7\times7$. The output of the last convolution layer is extracted to be input to Readout Net.  This model is also based on the standard architecture without skip connections, as shown in Figure \ref{fig:main-archs} (b).

\item {\bf ResNet} \cite{resnet} has up to 152 layers by using the residual learning framework, where Convolutions in Figure \ref{fig:main-archs} (a) is composed of identity mapping (skip connection) with the bypass connection to learn the residuals by the convolutions, as shown in Figure \ref{fig:main-archs} (c). In general, the training of CNN with deep layers is difficult because gradients can be easily vanished or exploded in the back propagation. Because of the direct connections for the identity mapping, the training of the deep layered neural networks becomes possible using the back propagation in ResNet. It is noted that the required parameters in ResNet are much smaller than VGG models in spite of the deeper layers such as 152 layers because of the small number of channels. The output of the last convolution layer is extracted to be input to Readout Net.

Let {\bf x} and  {\bf y} be input and output feature maps of a convolution layer in ResNet, respectively. A convolution layer of ResNet can be expressed as
\begin{equation}
{\bf y} = F({\bf x}) + {\bf x}
\end{equation}
where $F({\bf x})$ is a function to represent the residual mapping shown in Convolutions in Figure \ref{fig:main-archs} (c), which is implemented with the following structure:
\begin{equation}
F({\bf x})=BN(C_{1\times1}(\sigma(BN(C_{3\times3}(\sigma(BN(C_{1\times1}({\bf x}))))))))
\end{equation}
$BN$, $C_{n\times n}$, and $\sigma$ denote the batch normalization, the $n\times n$ convolution layer, and the activation function of ReLU, respectively. Note that only ResNet-18 has different $F$ as follows:
\begin{equation}
F({\bf x})=BN(C_{3\times3}(\sigma(BN(C_{3\times3}(\sigma({\bf x}))))))
\end{equation}

\item {\bf  SqueezeNet} \cite{squeezenet} with 18 layers has been proposed as a smaller CNN architecture compared with AlexNet, which is based on the standard architecture shown in Figure \ref{fig:main-archs} (b). Each 'Convolutions' in Figure \ref{fig:main-archs} (b), called 'Fire module', consists of a squeeze convolution layer with 1$\times$1 convolution filters, and an expand layer with both 1$\times$1 and 3$\times$3 filters. The output of the last Fire module is extracted to be input to Readout Net.

\item {\bf DenseNet} \cite{densenet, efficient-densenet} has up to 264 layers, by using skip connections similar to ResNet. In contrast to ResNet where the feature maps from Convolutions are added to the directly connected feature maps from the preceding layer (Figure \ref{fig:main-archs}c), DenseNet concatenates the feature maps from Convolutions with those from the preceding layer along channels (Figure \ref{fig:main-archs}d). This results in each layer of DenseNet receiving all the outputs of preceding layers because the input of each layer is directly transmitted to the next layer by the skip connection.

DenseNet mainly consists of multiple dense blocks composed of dense layers with dense connections shown in Figure \ref{fig:main-archs} (d). Let the $ l $-th dense layer in a dense block be $F_l $ and its output be $ {\bf x}_l $. Then, ${\bf x}_l $ can be expressed as follows:
\begin{equation}
{\bf x}_l = F_l([{\bf x}_0, {\bf x}_1, ...,{\bf x}_{l-1}])
\end{equation}
where $ [{\bf x} _ 0, {\bf x} _ 1, ..., {\bf x} _ {l - 1}] $ represents the concatenation of the feature maps $ {\bf x}_0 $ to $ {\bf x}_{l - 1} $. $ F_l $ is represented as 'Convolutions' in Figure \ref{fig:main-archs} (d), which is implemented with following structure:

\begin{equation}
F_l({\bf x})=C_{3\times3}(\sigma(BN(C_{1\times1}(\sigma(BN({\bf x}))))))
\end{equation}

Since the input of each dense layer has a large number of channels, BN-ReLU-$1\times1\,$Conv is inserted before the BN-ReLU-$3\times3\,$Conv to condense the input information by reducing the channels. 

The number of input channels in the $l$-th dense layer is $H+(l-1)K$ channels, where $H$ is the number of channels in the dense block input and $K$ is the number of channels in the $3\times3$ convolution output, called growth rate. The $1\times1$ convolution and the $3\times3$ convolution reduces the channels to $4K$ and $K$, respectively. The output of the $l$-th dense layer is the concatenation of the input ($H+(l-1)K$ channels) and the $3\times3$ convolution output ($K$ channels), resulting in the $H+l\times K$ channels. When a dense block is composed of $L$ dense layers, the output of the dense block has $H+KL$ channels. Each dense block except for the last one is followed by a transition layer which reduces channels using 1$\times$1 convolution and reduces the resolution of feature maps using average pooling.

\item {\bf DPN} \cite{dpn} used in the experiments has 131 layers, combining ResNet and DenseNet structures. As shown in Figure \ref{fig:main-archs} (e), the output of 'Convolutions' is divided into two parts: one is to be added with the directly connected feature maps like ResNet, and the other is to be concatenated with them like DenseNet.The number of channels in the input and output for a layer in DPN is same as that in DenseNet, as shown in Figure \ref{fig:main-archs} (d) and (e). The 'Convolutions' in DPN is composed of a $1\times1$ convolution layer followed by a $3\times3$ convolution layer, and another $1\times1$ convolution layer. In the $3\times3$ convolution layer, the grouped convolution is used as proposed in ResNeXt \cite{resnext}, where the feature maps are grouped along the channels before the convolution. This grouped convolution enhances the learning capacity and efficiency of the convolution. The output of the last convolution layer is extracted to be input to Readout Net.
\end{itemize}

\subsection{Components of Readout Net}\label{sec:methods:readout}

Readout Net examined is a multi-layer neural network composed of bilinear interpolation layers (BI), deconvolution layers (DC), or sub-pixel convolution layers (SPC), followed by a $1\times1$ convolution layer and Leaky ReLU. The last $1\times1$ convolution layer outputs the estimated saliency map using the concatenated feature maps from both two scales of Main Nets ($\times$1.0, $\times$0.5). 

\begin{itemize}
\setlength{\leftskip}{0.3cm}
\item {\bf Deconvolution (DC):} Deconvolution is also called as transposed convolution. The operation attempts to directly minimize the reconstruction error of the input image under a sparsity constraint on an over-complete set of feature maps \cite{deconvolution}. When DC is used for the upsampling layers in Readout Net, the 1st, 2nd, and 3rd DC followed by a ReLU layer in Readout Net reduce the channels to 128, to 64, and to 32, respectively. Then, the $1\times1$ convolution reduce the channel to 1 to predict the saliency map. The filter size of DC was set to $4\times4$.
\item {\bf Sub-Pixel Convolution (SPC):}  SPC \cite{spc} is a method to recover a high resolution image from its low resolution counter part with little additional computational cost, by rearranging the data along the channel into feature maps with a convolution operation. When SPC is used for the upsampling layers, each SPC layer reduces the channels to one forth, followed by a $3\times3$ convolution and a ReLU layer. Then, the $1\times1$ convolution predicts the saliency map from the output of the last SPC layer. 
\item {\bf Bilinear Interpolation (BI):} Since each BI layer in the upsampling network resize a feature map twice while maintaining the feature-map channels, GPU was out of memory when 3 upsampling layers were used for the all channels of outputs of Main Net. In the case of BI, the order of upsampling and projection ($1\times1$ convolution) can be inverted without any influence on the output. Therefore, the concatenated feature maps from Main Nets are first processed by the $1\times1$ convolution to output the 1-channel feature map, followed by the BI upsampling layers.
\end{itemize}

\section{Experimental setup}\label{sec:experimental_setup}

The datasets used in our experiments are explained. The network was fine-tuned on training data in Salicon Dataset and OSIE, except for the experiment for MIT Saliency Benchmark (Table \ref{tb:densesal-mit300}), where MIT1003 was also used for training. The datasets of PASCAL-S and MIT300 (MIT Saliency Benchmark) were used for the evaluation.

For most of the experiments, Pascal-S was used for the evaluation because the reference paper \cite{saliconnet} of the SaliconNet model also used this dataset as one of the evaluation datasets. The evaluation results for different databases in addition to Pascal-S are described in \ref{sec:results:state_of_the_art} for the comparison with the state-of-the-art models.

\begin{itemize}
\setlength{\leftskip}{0.3cm}
\item {\bf Salicon Dataset} \cite{salicondataset} consists of 10,000 images for training and 5,000 images for validation. Each image was viewed by 60 observers. Different from other fixation datasets, this dataset is largescale mouse-tracking data through Amazon Mechanical Turk. Although this dataset is not the fixation dataset, it is well known that the distribution of mouse tracking points is similar to the distribution of fixations \cite{salicondataset}, and that the parameters of the model trained on Salicon Dataset is useful for saliency map estimation \cite{deepgaze2, densesal}. The training data in the Salicon Dataset was used for training the network, whereas the validation data was used for evaluating models in \ref{sec:results:scatter-plot} (Figure \ref{fig:scatter-plot}).

\item {\bf OSIE} \cite{osie} consists of 700 natural indoor and outdoor scenes, aesthetic photographs from Flickr and Google. The 500 images were used for training and the 200 images for validation. The fixations were measured while 15 observers looked at image for 3 seconds. In order to obtain the optimal saliency maps from the fixations, the fixations for all observers were collected and blurred using the gaussian filter with the standard deviation equivalent to 1 degree in the visual angle. The training data in the OSIE dataset was used for training the network after the training with Salicon Dataset, whereas the validation data was used for evaluating the estimated saliency maps in Table \ref{tb:dpnsal-osie}.

\item {\bf PASCAL-S} \cite{pascals} is built on the subset of the validation data in the PASCAL VOC 2010 \cite{pascalvoc} segmentation challenge. PASCAL-S contains 850 natural images. The fixations were measured while 8 observers looked at an image for 2 seconds. This dataset was used for evaluating the estimated saliency maps to investigate the generalization capability of the models trained on OSIE dataset. Note that this dataset was not used for training. This procedure is same as \cite{saliconnet}.

\item {\bf MIT1003} \cite{mit1003} includes 779 landscape images and 228 portrait images. The fixations were measured while 15 observers looked at an image for 3 seconds. This dataset was used for the evaluation in Table \ref{tb:dpnsal-mit1003}. This dataset is also used as training dataset for MIT Saliency Benchmark in Table \ref{tb:densesal-mit300}, where 902 images were used for training and 101 images for validation. 

\item {\bf MIT300} \cite{mit300} was the first data set with held-out human eye movements, and is used as benchmark test data in MIT Saliency Benchmark in Table \ref{tb:densesal-mit300}. This dataset consists of 300 natural indoor and outdoor scenes. The fixations were measured while 39 observers looked at an image for 3 seconds.
\end{itemize}

Saliency map is evaluated in several metrics. Each metric is categorized as the location-based or distribution-based metric. These types depend on whether the ground truth is represented as discrete fixation locations or a continuous fixation map. Metrics used in our experiments are explained below, where lower values represent better performance of the saliency map estimation for Kullback-Leibler Divergence (KL) and Earth Mover's Distance (EMD), while higher values represent better performance for the other metrics.

\begin{itemize}
\setlength{\leftskip}{0.3cm}

\item {\bf AUC-Judd} is a location-based metric. The true positive rate (TPR) is calculated as the proportion of fixations falling into the thresholded saliency map \cite{metrics}. The false positive rate (FPR) is calculated as the proportion of no-fixated pixels in the thresholded saliency map. After calculating TPR and FPR at each threshold, Area Under the Curve (AUC) is calculated for the curve of TPR against FPR.
\item {\bf AUC-Borji} is a location-based metric. TPR is calculated in the same way as AUC-Judd \cite{metrics}. FPR is obtained by calculating the proportion of negatives in the thresholded region, where the negatives are collected uniformly at random. AUC for the curve is calculated as AUC-Borji.
\item {\bf shuffled-AUC (sAUC)} is a location-based metric. TPR is calculated in the same way as AUC-Judd \cite{metrics}. FPR is calculated based on the negatives which are determined by fixation points of all the other images in the dataset. AUC for the curve is calculated as sAUC. This metric can eliminate the center bias property of human fixations, which is the property that a human tends to look at the center of an image.
\item {\bf Kullback-Leibler Divergence (KL)} \cite{kld} is a distribution-based metric, which measures the difference between two probability density functions. This KL was used as the loss function to train the neural networks to compare our proposed models with SaliconNet, one of the competitive methods, which has used KL as the loss function.
\item {\bf Normalized Scanpath Salience (NSS)} \cite{nss} is a location-based metric. Each saliency map was normalized to have the zero mean and the unit standard deviation. The values of the normalized saliency map at fixations are averaged to evaluate the saliency map.
\item {\bf  Correlation Coefficient (CC)} \cite{cc} is a distribution-based metric, which is also called Pearson's linear coefficient. This metric is the linear correlation coefficient between the estimated saliency map and the optimal saliency map.
\item {\bf Similarity (SIM)} \cite{sim} is a distribution-based metric. This metric is computed as the sum of the minimum values of the estimated saliency map and the optimal saliency map for all pixels.
\item {\bf Earth Mover's Distance (EMD)} \cite{metrics} is a distribution-based metric. EMD measures the spatial distance between two probability density functions.
\end{itemize}

First, the influence of image classification accuracy on the saliency map estimation task was investigated in \ref{sec:results:scatter-plot}. The 15 models based on various architectures in Figure \ref{fig:main-archs} were tested in this experiment. These models are listed in Table \ref{tb:info-mainnet}. Since the computational cost was too high to evaluate all the models with the multipath architecture in Figure \ref{fig:arch}, the single-path structure for full-size images ($\times$ 1.0) was used with the 1 $\times$ 1 convolution as Readout Net in this experiment. The pretrained models that are provided by torchvision \cite{torchvision} were used. As for DenseNet-264 and DPN-131, we used pretrained models that are provided by the authors \cite{efficient-densenet, dpn}. Due to the high computational cost, the validation dataset of Salicon Dataset was used for the evaluation. Since Salicon Dataset was used for training the network, evaluation with the validation dataset of Salicon Dataset can suppress the evaluation time. DenseNet-161 and DPN-131 gave high accuracy in this experiment for the saliency map estimation, so that these two models were used in the following experiments.

\begin{table}[t]
  \begin{center}
    \caption{Comparison of models used as Main Net}
    \label{tb:info-mainnet}
    \centering
    \scalebox{0.99}{
      \begin{tabular}{|c|c c c c|} \hline
        Model & \# layers & \# parameters & Top 1 Acc & Top 5 Acc \\ \hline \hline
        AlexNet & 8 & 61.1 M  & 56.55 & 79.09 \\
        VGG & 11 & 132.9 M & 69.02 & 88.63 \\
        VGG & 13 & 133.0 M & 69.93 & 89.25 \\
        VGG & 16 & 138.4 M & 71.59 & 90.38 \\
        VGG & 19 & 143.7 M & 72.38 & 90.88 \\
        ResNet & 18 & 11.7 M & 69.76 & 89.08 \\
        ResNet & 50 & 25.6 M & 76.15 & 92.87 \\
        ResNet & 101 & 44.5 M & 77.37 &  93.56\\
        ResNet & 152 & 60.2 M & 78.31 & 94.06 \\
        SqueezeNet & 18  & 1.2 M & 58.10 & 80.42 \\
        DenseNet & 121 & 8.0 M & 74.65 & 92.17 \\ 
        DenseNet & 169 & 14.1 M & 76.00 & 93.00 \\
        DenseNet & 161 & 28.7 M & 77.65 & 93.80 \\
        DenseNet & 264 & 72.7 M & 79.74 & --- \\
        DPN & 131 & 79.3 M & 80.07 & 94.88 \\ \hline
      \end{tabular}
    }
  \end{center}
\end{table}

Using the two models (DenseNet-161 and DPN-131), the effective architecture and training procedure for saliency map estimation was studied with the structure of Figure \ref{fig:arch} in \ref{sec:results:architectures}-\ref{sec:results:examples}. In these experiments, the network was pretrained with the image classification task followed by fine-tuning with Salicon Dataset and OSIE. Then, Pascal-S was used for the evaluation. This procedure was same as the reference of the Salicon model \cite{saliconnet}. In Section \ref{sec:results:state_of_the_art}, the different databases were also used for the evaluation to compare the accuracy with the state-of-the-art models. The architectures using DenseNet-161 and DPN-131 with the structure of Figure \ref{fig:arch} are called DenseSal and DPNSal, respectively. The detailed structures are shown in Table \ref{tb:details-densesal} and \ref{tb:details-dpnsal}. 

The DenseNet-161 includes 4 dense blocks as shown in Table \ref{tb:details-densesal}. Although the original DenseNet-161 has the average pooling layer with the stride 2 between the 3rd and 4th dense blocks as in the other pooling layers, the stride was set to 1 at the layer to predict the saliency map in the higher resolution. The feature maps before the final layer in DenseNet-161 (Global Pooling in the original DenseNet-161) are used as the output of Main Net in the proposed model. The output size of the Main Net is the 1/16 of the input image, whereas it is the 1/32 in the original DenseNet-161. The details of the DenseSal structure are shown in Table \ref{tb:details-densesal}.

The DPN-131 was also modified to have higher resolution in the output by adjusting the stride in the $3\times3$ convolution layer in the 4th block, as shown in Table \ref{tb:details-dpnsal}. In this model, the dilated convolution \cite{deeplab} was employed to enlarge the receptive filed in the 4th block. The output size of the Main Net is the 1/16 of the input image, whereas it is the 1/32 in the original DPN-131.

KL and RMSprop \cite{rmsprop} were used as the objective function and as the solver, respectively. PyTorch \cite{pytorch} was used as the implementation platform. The mini-batch size and learning rate were set to 1 and $10^{-5}$ during training, respectively. We subtract the per-channels mean value of training images from each image as pre-processing. Note that the 2015 version of Salicon Dataset was used in the experiments except for the experiments in Figure \ref{fig:scatter-plot}, where the 2017 version of Salicon Dataset was used. Since it was found that the performance of the model trained on the 2015 version of the Salicon Dataset was better than that on the 2017 version, we decided to use the 2015 version in most of the experiments. 

\begin{table}[t]
  \begin{center}
    \caption{Details of the model based on DenseNet-161: DenseSal \newline (N: Number of upsampling layers in Readout Net, \newline The symbol (+K): the width increment on the densely connected path)}
    \label{tb:details-densesal}
    \centering
    \scalebox{0.78}{
      \begin{tabular}{ c | c | c c c c } \hline
        && \multicolumn{2}{c}{Main Net ($\times1.0$)} &  \multicolumn{2}{c}{Main Net ($\times0.5$)}  \\ \cmidrule(r){1-6} 
        \multirow{2}{*}{Stage} & \multirow{2}{*}{Filter size} & \multirow{2}{*}{\footnotesize{Channels}} & \footnotesize{Output} & \multirow{2}{*}{\footnotesize{Channels}} & \footnotesize{Output} \\
        &&&\footnotesize{size}&&\footnotesize{size} \\ \cmidrule(r){1-6} \cmidrule(r){1-6}

        Conv & $7\times7 \: (stride=2)$ & 96 & 1/2 & 96 & 1/4 \\\cmidrule(r){1-6}
        Pooling &$3\times3 \: (stride=2)$ & 96 & 1/4 & 96 & 1/8\\\cmidrule(r){1-6}
        Dense Block & L=6 (+48) & 384 & 1/4 & 384 &1/8 \\\cmidrule(r){1-6}
        Conv &$1\times1\:(stride=1)$ & 192 & 1/4 & 192 & 1/8 \\\cmidrule(r){1-6}
        Pooling &$2\times2\: (stride=2)$ & 192 & 1/8 & 192 & 1/16\\\cmidrule(r){1-6}
        Dense Block &L=12 (+48) & 768 &1/8 & 768 &1/16 \\\cmidrule(r){1-6}
        Conv &$1\times1\: (stride=1)$ & 384 & 1/8 & 384 & 1/16 \\\cmidrule(r){1-6}
        Pooling &$2\times2\: (stride=2)$ & 384 & 1/16 & 384 & 1/32\\\cmidrule(r){1-6}
        Dense Block & L=36 (+48) & 2112 & 1/16 & 2112 & 1/32 \\\cmidrule(r){1-6}
        Conv &$1\times1\: (stride=1)$ & 1056 & 1/16 & 1056 & 1/32 \\\cmidrule(r){1-6}
        Pooling &$2\times2\: (stride=1)$ & 1056 & 1/16 & 1056 & 1/32\\\cmidrule(r){1-6}
        Dense Block & L=24 (+48) & 2208 & 1/16 & 2208 & 1/32 \\\cmidrule(r){1-6}
        \raisebox{0.7em}{Concatenation} & & \raisebox{0.7em}{4016} & \raisebox{0.7em}{1/16} & \multicolumn{2}{c}{\huge{$\gets$}} \\ \cmidrule(r){1-6}
        Readout Net & &  1 & $2^N$/16 & \multicolumn{2}{c}{} \\ \cmidrule(r){1-6}
      \end{tabular}
    }
  \end{center}
\end{table}

\begin{table}[t]
  \begin{center}
    \caption{Details of the model based on DPN-131: DPNSal\newline (N: Number of upsampling layers in Readout Net, \newline The symbol (+K): the width increment on the densely connected path)}
    \label{tb:details-dpnsal}
    \centering
    \scalebox{0.75}{
      \begin{tabular}{ c | c | c c c c } \hline
        && \multicolumn{2}{c}{Main Net ($\times1.0$)} &  \multicolumn{2}{c}{Main Net ($\times0.5$)}  \\ \cmidrule(r){1-6} 
        \multirow{2}{*}{Stage} & \multirow{2}{*}{Filter size} & \multirow{2}{*}{\footnotesize{Channels}} & \footnotesize{Output} & \multirow{2}{*}{\footnotesize{Channels}} & \footnotesize{Output} \\
        &&&\footnotesize{size}&&\footnotesize{size} \\ \cmidrule(r){1-6} \cmidrule(r){1-6}
        Conv & $7\times7\: (stride=2)$ & 128 & 1/2 & 128 & 1/4 \\ \cmidrule(r){1-6}
        &$3\times3\:max pool\: (stride=2)$ & & & & \\ \cmidrule(r){2-2}
        \raisebox{1em}{Block1} & $\begin{bmatrix} 1\times1, \\ 3\times3\:(stride=1) \\ 1\times1\:(+16)  \end{bmatrix} \times 4$ & \raisebox{-1em}{352} & \raisebox{-1em}{1/4} & \raisebox{-1em}{352} &\raisebox{-1em}{1/8} \\ \cmidrule(r){1-6}
        \multirow{5}{*}{Block2}& $\begin{bmatrix} 1\times1 \\ 3\times3\: (stride=2) \\ 1\times1\:(+32)  \end{bmatrix} \times 1$ &  &  &  & \\ \cmidrule(r){2-2}
        & $\begin{bmatrix} 1\times1 \\ 3\times3\:(stride=1) \\ 1\times1\:(+32)  \end{bmatrix} \times 7$ & \raisebox{-1em}{832} & \raisebox{-1em}{1/8} & \raisebox{-1em}{832} & \raisebox{-1em}{1/16} \\ \cmidrule(r){1-6}
        \multirow{5}{*}{Block3} & $\begin{bmatrix} 1\times1 \\ 3\times3\:(stride=2) \\ 1\times1\:(+32)  \end{bmatrix} \times 1 $ &  &  &  & \\ \cmidrule(r){2-2}
        & $\begin{bmatrix} 1\times1 \\ 3\times3\:(stride=1) \\ 1\times1\:(+32)  \end{bmatrix} \times 27 $ & \raisebox{-1em}{1984} & \raisebox{-1em}{1/16} & \raisebox{-1em}{1984} & \raisebox{-1em}{1/32} \\ \cmidrule(r){1-6}
        Block4 & $\begin{bmatrix} 1\times1 \\ 3\times3\:(stride=1) \\ 1\times1\:(+128) \end{bmatrix} \times 3 $ & \raisebox{-1em}{2688} & \raisebox{-1em}{1/16} & \raisebox{-1em}{2688} & \raisebox{-1em}{1/32} \\ \cmidrule(r){1-6}
        \raisebox{0.7em}{Concatenation} & & \raisebox{0.7em}{5376} & \raisebox{0.7em}{1/16} & \multicolumn{2}{c}{\huge{$\gets$}} \\ \cmidrule(r){1-6}
        Readout Net & &  1 & $2^N$/16 & \multicolumn{2}{c}{} \\ \cmidrule(r){1-6}
      \end{tabular}
    }
  \end{center}
\end{table}

\section{Experimental results}\label{sec:results}

\subsection{Influence of image classification accuracy on saliency map estimation}\label{sec:results:scatter-plot}

\begin{figure*}[t]
  \begin{center}
    \includegraphics[scale=0.45]{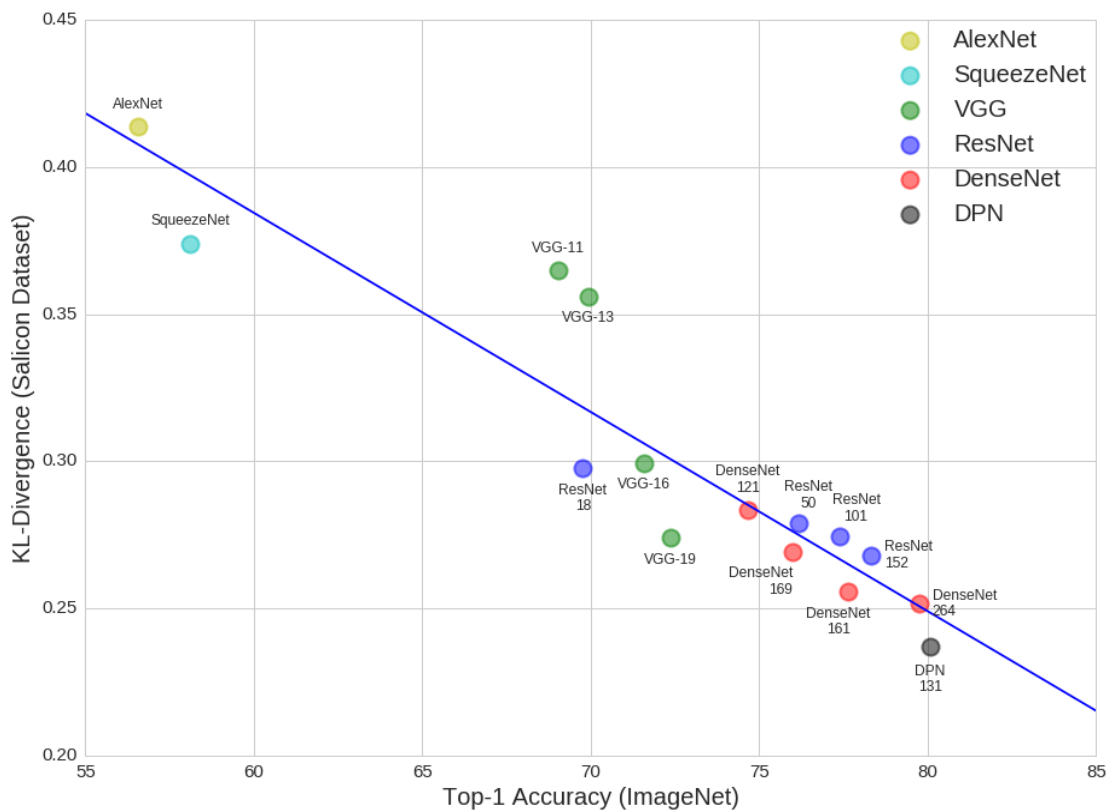}
  \end{center}
  \caption{Saliency-map estimation accuracy (KL) against image classification accuracy (Top-1 Accuracy). Pearson correlation coefficient: -0.927 (p-value < 0.05)}
  \label{fig:scatter-plot}
\end{figure*}

The 15 models shown in Table \ref{tb:info-mainnet} were examined to study the relationship of the saliency map estimation accuracy with the image classification accuracy. The single-path architecture with 1 $\times$ 1 convolution as Readout Net was used in this experiment to suppress the computational cost. Main Net of the each model was initialized with the weights trained on ImageNet, followed by fine-tuning with the training dataset of Salicon Dataset. The validation dataset of Salicon Dataset was used for the evaluation in this experiment to suppress the computational cost. The learning rate was set to $10^{-5}$ for first 10 epochs and $10^{-6}$ next 10 epochs in this experiment. It is shown in Figure \ref{fig:scatter-plot} that there is a strong correlation between the image classification accuracy on ImageNet and the saliency-map estimation accuracy evaluated by KL on the validation data in Salicon Dataset. The Pearson correlation coefficient is $-0.927$ with statistically significant value (p < 0.05). Similar results were observed for other metrics: NSS, CC, and SIM. Since it takes time to calculate AUC-based metrics and EMD for the databases with large number of fixations, only the four metrics were examined in this experiment Among the 15 models, the model with DPN-131 achieved the best performance in all metrics. It is seen from these results that the ability of object recognition is important for saliency map estimation.

\subsection{Comparison of architectures}\label{sec:results:architectures}

In the following experiments (\ref{sec:results:architectures}-\ref{sec:results:examples}), DenseNet-161 and DPN-131 were used with the structure of Figure \ref{fig:arch} (DenseSal and DPNSal). VGG-16 with the structure of Figure \ref{fig:arch} (SaliconNet model \cite{saliconnet} with N=0 in Readout Net) was used as the baseline. In the default settings, the networks pretrained with ImageNet were fine-tuned with Salicon Dataset and OSIE. Pascal-S was used for the evaluation, as in \cite{saliconnet}.

First, VGG-16, DenseNet-161, and DPN-131 were compared using a $ 1 \times 1 $ convolution layer as Readout Net (N=0). The sizes of outputs in these models are the 1/16 of the input images. The model using VGG-16 is the same as SaliconNet. 
In Table \ref{tb:multipath}, these models were compared by evaluated with KL on the PASCAL-S dataset, though all the results for the other metrics are also provided in Appendix. The KL metrics was selected to be shown in the text since the metric was used as the objective function during the training.

The KL values of DenseNet-161 and DPN-131 with the multipath structure were lower than VGG-16, meaning that the accuracy of estimating saliency maps was improved by using DenseNet-161 and DPN-131, while the accuracy of DenseNet-161 and DPN-131 were same. In addition, it was confirmed that the accuracy was improved by using two scales ($\times 1.0$ and $\times 0.5$) in these models.

In addition to the accuracy, the computational costs were compared among these models, as shown in Table \ref{tb:compcost}. As can be seen from the table, the computational costs for DenseNet and DPN were about 2.5 times and 5.2 times of VGG, respectively. The cost in the single path with the half-size image was about the half of that in single path with the full-size image, while the cost in multipath was about 1.4 times of the single path with the full-size image. Thus, there was a trade-off between the accuracy and the computational cost.

\begin{figure*}[t]
  \begin{center}
    \includegraphics[width=18cm, bb=0 0 1286 494]{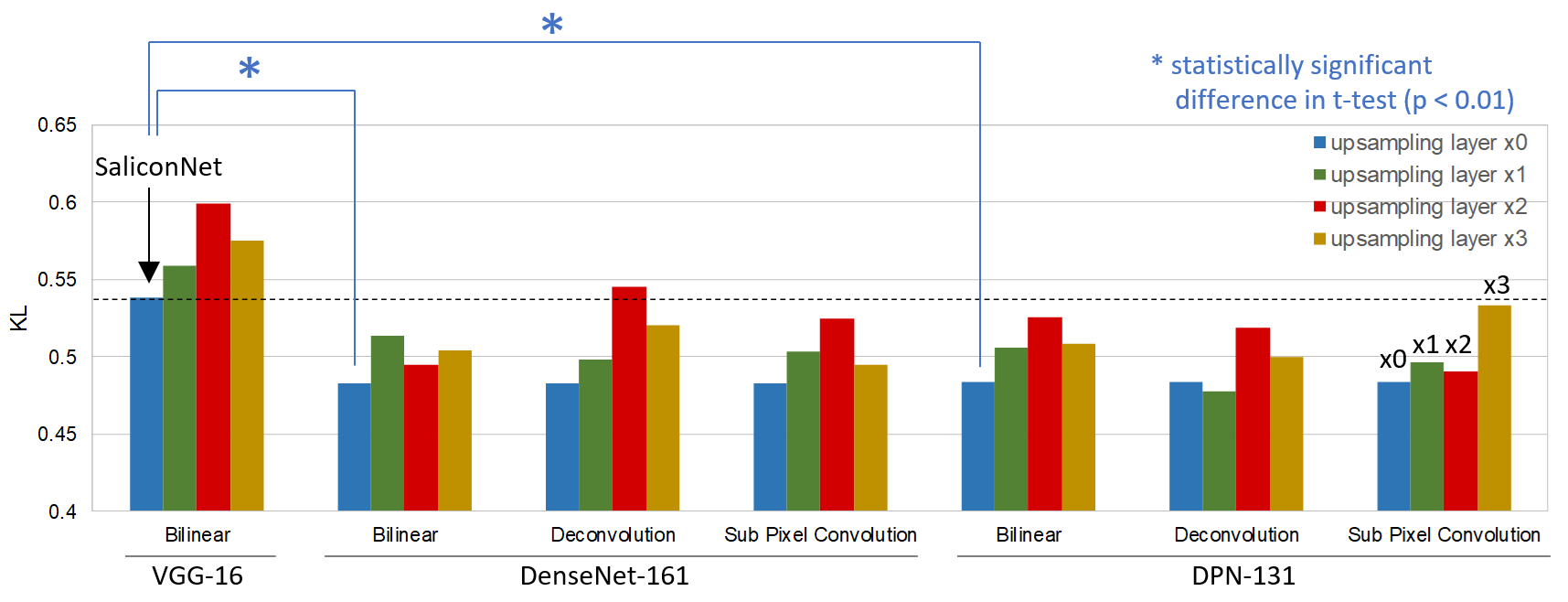}
  \end{center}
  \caption{Comparison of upsampling-layer types in Readout Net on PASCAL-S}
  \label{fig:upsamplings}
\end{figure*}

\begin{table}[t]
  \begin{center}
    \caption{Comparison of network architectures (KL on PASCAL-S)}
    \label{tb:multipath}
    \centering
    \scalebox{1.0}{
      \begin{tabular}{|l|c c c|} \hline
        Architecture & VGG-16 & DenseNet-161 & DPN-131 \\ \hline \hline
        Single path ($\times$0.5) & 0.636 & 0.507 & 0.522 \\
        Single path ($\times$1.0) & 0.593 & 0.511 & 0.502 \\ 
        Multi path ($\times$1.0, $\times$0.5) & 0.538 & {\bf 0.483} & {\bf 0.483} \\\hline
      \end{tabular}
    }
  \end{center}
\end{table}

\begin{table}[t]
  \begin{center}
    \caption{Comparison of computational cost (ms/image)}
    \label{tb:compcost}
    \centering
    \scalebox{0.95}{
      \begin{tabular}{|c l |c c c|} \hline
        Architecture & & VGG-16 & DenseNet-161 & DPN-131 \\ \hline \hline
        Single path & training & 27.9 & 83.2 & 200.1 \\
        ($\times$0.5) & prediction & 7.4 & 23.9 & 42.2 \\
        \hline
        Single path & training & 75.6 & 146.9 & 336.5 \\
        ($\times$1.0) & prediction & 19.4 & 42.7 & 76.2 \\
        \hline
        Multi path & training & 93.9 & 216.5 & 512.5 \\
        ($\times$1.0, $\times$0.5) & prediction & 25.5 & 64.8 & 116.2 \\
        \hline
      \end{tabular}
    }
  \end{center}
\end{table}

\begin{table}[t]
  \begin{center}
    \caption{Comparison of training datasets (KL on PASCAL-S)}
    \label{tb:training-dataset}
    \centering
    \scalebox{1.0}{
      \begin{tabular}{|l|c c c|} \hline
        \multirow{2}{*}{Training Datasets} & VGG & DenseNet & DPN \\
        & -16 & -161 & -131 \\ \hline \hline
        OSIE & 0.845 & 1.225 & 0.966\\
        Salicon $\to$ OSIE & 0.686 & 0.814 & 0.842 \\
        ImageNet $\to$ OSIE & 0.648 & 0.632 & 0.565\\
        ImageNet $\to$ Salicon $\to$ OSIE & 0.538 & {\bf 0.483} & {\bf 0.483} \\ \hline
      \end{tabular}
    }
  \end{center}
\end{table}

\subsection{Comparison of training datasets}\label{sec:results:training_datasets}

In CNN, the overfitting to the training data can be suppressed by using the initialization of the network parameters trained with a large amount of image data such as ImageNet. For the saliency map estimation, it is known that the parameters of the model trained on ImageNet for image classification and Salicon Dataset are useful \cite{saliconnet, deepgaze2}. Therefore, the influence of datasets used for training the network was investigated. The evaluation results with KL on Pascal-S after the training is shown in Table \ref{tb:training-dataset} for the model with the multipath architecture and a $ 1 \times 1 $ convolution layer as Readout Net (see Appendix for the other metrics). From the results in Table \ref{tb:training-dataset}, it is clear that the model parameters trained on ImageNet were most important. Furthermore, the training with Salicon Dataset slightly improved the accuracy. 

It is also shown that the models with DenseNet-161 and DPN-131 achieved better performance than the model with VGG-16 when the ImageNet was used for the training. On the contrary, their performances were worse than VGG-16 without the ImageNet training. The reason of these results would be overfitting to the training data in DenseNet-161 and DPN-131 with small training dataset such as OSIE. From these results, not only the architecture of the networks but also the parameters pretrained with the ImageNet classification task are important for the saliency map estimation.

In the following experiments, the models were trained by ImageNet, Salicon Dataset, and OSIE, in order unless otherwise stated. During the training with OSIE, the input images were resized to 640$\times$480 due to the GPU memory limitation.

\subsection{Comparison of upsampling layers in Readout Net}\label{sec:results:readoutnet}

Readout Net aims to convert the features extracted by Main Net into a saliency map. Since the output size of Main Net is much smaller than the original input image, the 4 types of upsampling neural networks were examined to estimate the saliency map in the higher resolution. Figure \ref{fig:upsamplings} is the results on the PASCAL-S evaluated by KL used as the objective function. The results for the other metrics are shown in Appendix.

It can be seen from Figure \ref{fig:upsamplings} that most of the results for DenseSal and DPNSal showed higher accuracy than the results on SalinonNet (VGG-16, N=0) shown in the doted lines on the graph. The accuracy on DenseSal and DPNSal was not improved as the upsampling layers increased. This result indicates that the upsampling layer is not necessary in the saliency map estimation task. Different from semantic segmentation or instance segmentation, the fine prediction is not important because the saliency map is blurred by convolving ground truth fixation points with a gaussian filter. In addition, the type of the upsampling networks did not affect the accuracy, as can be seen from the graph.

\subsection{Comparison with state-of-the-art models}\label{sec:results:state_of_the_art}

The results of DenseSal and DPNSal with the $1\times1$ convolution for Readout Net were compared with the conventional methods on various evaluation metrics on PASCAL-S, as shown in Table \ref{tb:densesal-pascals}. ITTI \cite{itti}, GBVS \cite{gbvs}, SUN \cite{sun}, DVA \cite{dva}, SIG \cite{sig}, and AWS \cite{aws} are the methods using manually designed image features. SaliconNet and DeepGase\,I\hspace{-.1em}I based on CNN showed the high accuracy in the results, where SaliconNet was evaluated for both authors' implementation through the webtool \cite{saliconnet} and our implementation (VGG-16, N=0). As can be seen from the table that DenseSal or DPNSal achieved the state-of-the-art accuracy on all the evaluation metrics. Thus, it was confirmed that DenseNet and DPN were useful not only for image classification but also for the saliency map estimation.

In order to compare the performance of our models with the state of the art methods, the saliency maps for the MIT300 were submitted to MIT Saliency Benchmark \cite{mit300}. DenseSal and DPNSal with the initialization of weights by the ImageNet classification task were fine-tuned with the datasets of Salicon Dataset, OSIE, and MIT1003, in order. The images in MIT1003 and MIT300 were resized smaller than or equal to $640\times480$ pixels or $480\times640$ pixels. It can be seen from Table \ref{tb:densesal-mit300} that DenseSal achieved the best performance in KL and EMD, whereas the competitive results in the other metrics. DPNSal also achieved the best performance in SIM, sAUC, CC, and NSS, with competitive results in the other metircs.

Furthermore, DenseSal and DPNSal were compared with SaliconNet \cite{saliconnet} on the validation data in OSIE and MIT1003. In these experiments, the networks were fine-tuned using Salicon Dataset and the training dataset of OSIE with the initialization of the ImageNet classification task, so that MIT1003 was untrained database. Only the four metrics (SIM, CC, NSS, KL) were used for the evaluation because it takes time to calculate AUC-based metrics and EMD for these databases with the large number of fixations. The results are shown in Table \ref{tb:dpnsal-osie} and \ref{tb:dpnsal-mit1003}. It can be seen from the results that DenseSal and DPNSal outperformed over SaliconNet by a large margin. In these databases, DPNSal were better than DenseSal in all the metrics examined.

\begin{table*}[t]
  \begin{center}
    \caption{Evaluation results on PASCAL-S}
    \label{tb:densesal-pascals}
    \centering
    \scalebox{0.99}{
      \begin{tabular}{| c | l |c c c c c c c c|} \hline
        \multicolumn{2}{|c|}{Model} & AUC-Judd $\uparrow$ & SIM $\uparrow$ & EMD $\downarrow$ & AUC-Borji $\uparrow$ & sAUC $\uparrow$ & CC $\uparrow$ & NSS $\uparrow$ & KL $\downarrow$\\ \hline \hline
        \multicolumn{2}{|c|}{Humans} & 0.967 & 1 & 0 & 0.979 & 0.848 & 1 & 3.906 & 0 \\ \hline
        & ITTI \cite{itti} & 0.845 & 0.419 & 2.262 & 0.853 & 0.610 & 0.517 & 1.394 & 1.053 \\
        & AIM \cite{aim} & 0.837 & 0.376 & 2.521 & 0.845 & 0.646 & 0.466 & 1.241 & 1.197 \\
        & GBVS \cite{gbvs} & 0.857 & 0.417 & 2.192 & 0.863 & 0.620 & 0.530 & 1.414 & 1.052 \\
        & SUN \cite{sun}  & 0.684 & 0.324 & 3.015 & 0.691 & 0.618 & 0.258 & 0.730 & 1.437 \\
        & DVA \cite{dva}  & 0.736 & 0.375 & 2.576 & 0.743 & 0.619 & 0.348 & 0.967 & 1.294 \\
        Conventional Methods& SIG \cite{sig}  & 0.753 & 0.358 & 2.731 & 0.753 & 0.643 & 0.374 & 1.067 & 1.293 \\
        & AWS \cite{aws} & 0.763 & 0.379 & 2.722 & 0.773 & 0.654 & 0.393 & 1.135 & 1.225 \\
        & Salicon \cite{saliconnet}  & 0.887 & 0.617 & 1.228 & 0.894 & 0.719 & 0.744 & 2.330 & 0.550 \\
        & Salicon(implemented)  & 0.891 & 0.620 & 1.268 & 0.901 & 0.733 & 0.759 & 2.390 & 0.538 \\
        & DeepGaze\,I\hspace{-.1em}I \cite{deepgaze2} & 0.859 & 0.573 & 1.664 & 0.864 & 0.723 & 0.671 & 2.130 & 0.776 \\ \hline
        \multirow{2}{*}{Proposed Methods} & DenseSal & 0.897 & 0.647 & 1.192 & {\bf 0.904} & {\bf 0.739} & {\bf 0.800} & 2.584 & {\bf 0.483} \\ 
        & DPNSal & {\bf 0.898} & {\bf 0.654} & {\bf 1.162} & {\bf 0.904} & 0.737 & 0.799 & {\bf 2.591} & {\bf 0.483} \\ 
        \hline
      \end{tabular}
    }
  \end{center}
\end{table*}

\begin{table*}[t]
  \begin{center}
    \caption{Evaluation results on MIT Saliency Benchmark (MIT300)}
    \label{tb:densesal-mit300}
    \centering
    \scalebox{1.0}{
      \begin{tabular}{| c | c c c c c c c c |} \hline
        Model & AUC-Judd $\uparrow$ & SIM $\uparrow$ & EMD $\downarrow$ & AUC-Borji $\uparrow$ & sAUC $\uparrow$ & CC $\uparrow$ & NSS $\uparrow$ & KL $\downarrow$\\ \hline \hline
        DeepFix \cite{deepfix} & 0.87 & 0.67 & 2.04 & 0.80 & 0.71 & 0.78 & 2.26 & 0.63 \\ 
        Salicon \cite{saliconnet} & 0.87 & 0.60 & 2.62 & 0.85 & {\bf 0.74} & 0.74 & 2.12 & 0.54 \\

        DeepGaze\,I\hspace{-.1em}I \cite{deepgaze2} & {\bf 0.88} & 0.46 & 3.98 & {\bf 0.86} & 0.72 & 0.52 & 1.29 & 0.96 \\
        SAM-ResNet \cite{samresnet} & 0.87 & 0.68 & 2.15 & 0.78 & 0.70 & 0.78 & 2.34 & 1.27 \\
        DSCLRCN \cite{dsclrcn} & 0.87 & 0.68 & 2.17 & 0.79 & 0.72 & 0.80 & 2.35 & 0.95 \\ \hline

        DenseSal & 0.87 & 0.67 & {\bf 1.99} & 0.81 & 0.72 & 0.79 & 2.25 & {\bf 0.48} \\
        DPNSal & 0.87 & {\bf 0.69} & 2.05 & 0.80 & {\bf 0.74} & {\bf 0.82} & {\bf 2.41} & 0.91 \\
        \hline
      \end{tabular}
    }
  \end{center}
\end{table*}

\begin{table}[t]
  \begin{center}
    \caption{Evaluation results on OSIE}
    \label{tb:dpnsal-osie}
    \centering
    \scalebox{1.0}{
      \begin{tabular}{| c | c c c c |} \hline
        Model   & SIM $\uparrow$ & CC $\uparrow$ & NSS $\uparrow$ & KL $\downarrow$\\ \hline \hline
        Salicon(implemented)  & 0.605 & 0.762 & 2.762  & 0.545\\ 
        DenseSal  & 0.659 & 0.822 & 3.068 & 0.443\\
        DPNSal  & \bf{0.686} & \bf{0.838} & \bf{3.175} & \bf{0.397}\\
        \hline
      \end{tabular}
    }
  \end{center}
\end{table}

\begin{table}[t]
  \begin{center}
    \caption{Evaluation results on MIT1003}
    \label{tb:dpnsal-mit1003}
    \centering
    \scalebox{1.0}{
      \begin{tabular}{| c | c c c c |} \hline
        Model   & SIM $\uparrow$ & CC $\uparrow$ & NSS $\uparrow$ & KL $\downarrow$\\ \hline \hline
        Salicon(implemented)  & 0.604 & 0.682 & 2.158  & 0.550\\ 
        DenseSal  & 0.641 & 0.751 & 2.439 & 0.467\\
        DPNSal  & \bf{0.692} & \bf{0.813} & \bf{2.678} & \bf{0.368}\\
        \hline
      \end{tabular}
    }
  \end{center}
\end{table}

\subsection{Examples of saliency map estimation}\label{sec:results:examples}

\begin{figure*}[t]
  \begin{center}
    \includegraphics[scale=0.532]{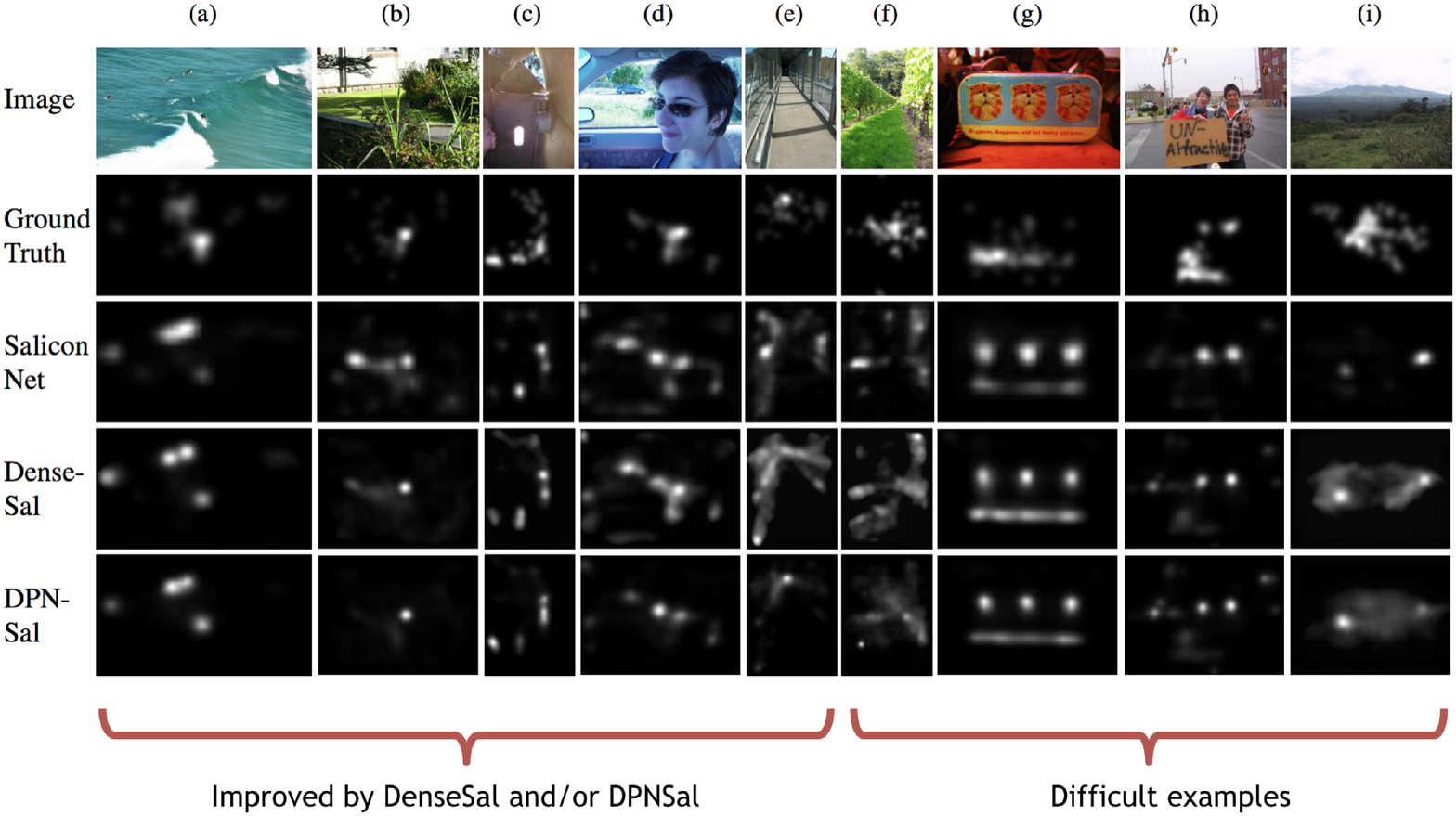}
  \end{center}
  \caption{Examples of saliency map estimation}
  \label{fig:examples}
\end{figure*}

The examples of saliency map estimation are shown in Figure \ref{fig:examples}, where (a)-(e) are the examples improved by DenseSal and/or DPNSal, while (f)-(i) are difficult examples. The improvement in (a)-(e) would be caused by better performance in the object recognition using DenseNet and DPN. 

In (a), the surfer in the center of the image is focused on by the observers. Although SaliconNet cannot estimate this area with high probability, DenseSal and DPNSal can capture this information. In (b), a squirrel is in the center of the image and is focused on by observers. Estimated probability well concentrated on the squirrel in the order of DPNSal, DenseSal, SaliconNet. DPNSal can capture the baby face in the image corresponding to the ground truth fixation map in (c), while SaliconNet cannot capture this. In (d), although the activations were observed in the saliency maps by SaliconNet and DenseSal at both the face and the small car in the background, DPN can focus on the face as the actual fixations. In (e), the human fixations concentrated on the end of passages. Although SaliconNet and DenseSal were unable to capture this information, DPN can do that. This result implies that DPN can learn the information even though there are few images in the training dataset. However, there are some example images of the passages which are not able to be estimated even with DPN, such as (f).

In (g), the actual fixations mainly concentrated on the texts, while the activations were observed in the saliency maps at the animal illustrations. To solve this problem, the model must consider the relationship and priority of fixations among objects in an image. In (h), the actual fixations also mainly concentrated on the texts. However, all models cannot capture that. The reason would be the number of images including handwritten characters is insufficient in the training dataset. In natural images without eye-catching objects such as (i), human fixations are often gathered in the center of an image, while the networks do not model this center-bias property.

\section{Conclusions}\label{sec:conclusions}

In this paper, the effective architecture was investigated for saliency map estimation, in addition to the influence of the image classification accuracy on the saliency map estimation.

First, it is shown that there is a strong correlation between image classification accuracy and saliency map estimation accuracy. In addition, it was found that not only the architecture but also the initialization strategy using the weights pretrained with the ImageNet classification task were important for estimating the saliency maps. These results indicate that the model which is pretrained with the ImageNet classification and has achieved high performance on the classification task is also useful for the saliency map estimation task. The reason for this would be that human fixations often concentrate on objects in image, while the model pretrained on ImageNet can react on many objects in images because ImageNet has a wide variety of object categories. If the model is initialized with random weights and is trained on a fixation dataset with the limited categories of objects for saliency map estimation, the model would overfit to the objects in the training dataset. On the contrary, if the model is trained for the image classification task which includes a wide variety of categories, overfitting for the objects in the training dataset would be suppressed owing to the large number of categories. Therefore, if the model trained on a large fixation dataset including a lot of objects such as ImageNet, it would be possible to achieve good performance for saliency map estimation without pretraining on ImageNet for the image classification task.

Second, the effective architecture and training procedure for saliency map estimation was investigated using DenseSal and DPNSal. It was confirmed that the multipath architecture was useful because the network was able to be robust to the size of objects in images. Moreover, it was found that the upsampling layers to refine the resolution of saliency maps were not necessary for the saliency map estimation because the saliency maps were created by blurring the fixation points. DenseSal and DPNSal achieved the state-of-the-art accuracy on PASCAL-S. Moreover, the models achieved the best performance in MIT Saliency Benchmark in most of the metrics.

In the future, the architecture that can find more complex structures should be studied for further improvement of saliency map estimation accuracy. From the results in this paper, the higher accuracy a model achieved in the image classification task, the higher saliency map estimation accuracy is expected in the saliency map estimation task using the model. However, it is thought that there is a limit to accuracy using such model. For example, the human fixations tend to concentrate on objects or regions which are looked at by a person in images \cite{wherenext}. Since this tendency is more complex than the tendency that the objects are often looked at by a human, the conventional methods including our model cannot estimate such information. To solve this problem, the model must estimate the gaze direction of a human in an image and capture objects or regions which are looked at by the person. Objective functions for training the models for saliency map estimation also need to be investigated. It is known that the metrics measure qualitatively different things. Therefore, if we use a metric as loss function, we will get better results in the metric but worse results in other metrics. This is why loss functions and metrics for saliency map estimation need to be studied more. Another future work may include the development of a network architecture in consideration of the center-bias property of human fixations \cite{deepgaze2}.



\section{Appendix}\label{sec:appendix}

The data for all the metrics for Table \ref{tb:multipath}, Table \ref{tb:training-dataset}, and Figure \ref{fig:upsamplings} are shown in Table \ref{tb:multipath-all}, \ref{tb:training-dataset-all}, and \ref{tb:upsamplings-all}, respectively. The properties discussed in the text are same for the other metrics. KL was selected to be shown in the text because this metric was used as the objective function during the training.

\begin{table*}[b]
  \caption{All the metrics for Table \ref{tb:multipath}, Table \ref{tb:training-dataset}, and Figure \ref{fig:upsamplings}}
  \label{tb:appendix:all-data}
  \subcaption{All the metrics for Table \ref{tb:multipath}: Comparison of network architectures}
  \label{tb:multipath-all}
  \scalebox{0.99}{
    \begin{tabular}{| c | l |c c c c c c c c|} \hline
      \multicolumn{2}{|c|}{Architecture} & AUC-Judd $\uparrow$ & SIM $\uparrow$ & EMD $\downarrow$ & AUC-Borji $\uparrow$ & sAUC $\uparrow$ & CC $\uparrow$ & NSS $\uparrow$ & KL $\downarrow$\\ \hline \hline
      \multirow{3}{*}{VGG-16} & Single path ($\times$0.5) & 0.878 & 0.571 & 1.423 & 0.888 & 0.716 & 0.693 & 2.041 & 0.636 \\
      & Single path ($\times$1.0) & 0.886 & 0.593 & 1.356 & 0.894 & 0.732 & 0.737 & 2.332 & 0.593\\
      & Multi path ($\times$1.0, $\times$0.5) & 0.891 & 0.620 & 1.268 & 0.901 & 0.733 & 0.759 & 2.399 & 0.538\\
      \hline
      \multirow{3}{*}{DenseNet-161} & Single path ($\times$0.5) & 0.894 & 0.631 & 1.225 & 0.902 & 0.736 & 0.785 & 2.495 & 0.507 \\
      & Single path ($\times$1.0) & 0.893 & 0.639 & 1.230 & 0.900 & 0.740 & 0.784 & 2.558 & 0.511\\
      & Multi path ($\times$1.0, $\times$0.5) & 0.897 & 0.647 & 1.192 & 0.904 & 0.739 & 0.800 & 2.584 & 0.483\\
      \hline
      \multirow{3}{*}{DPN-131} & Single path ($\times$0.5) & 0.893 & 0.628 & 1.242 & 0.901 & 0.733 & 0.774 & 2.460 & 0.522\\
      & Single path ($\times$1.0) & 0.896 & 0.645 & 1.208 & 0.904 & 0.736 & 0.784 & 2.536 & 0.502\\
      & Multi path ($\times$1.0, $\times$0.5) & 0.898 & 0.654 & 1.162 & 0.904 & 0.737 & 0.799 & 2.591 & 0.483\\
      \hline
    \end{tabular}
  }
  \newline \newline
  \subcaption{All the metrics for Table \ref{tb:training-dataset}: Comparison of training datasets}
  \label{tb:training-dataset-all}
  \scalebox{0.99}{
    \begin{tabular}{| c | l |c c c c c c c c|} \hline
      Model & Training Datasets & AUC-Judd $\uparrow$ & SIM $\uparrow$ & EMD $\downarrow$ & AUC-Borji $\uparrow$ & sAUC $\uparrow$ & CC $\uparrow$ & NSS $\uparrow$ & KL $\downarrow$\\ \hline \hline
      \multirow{4}{*}{VGG-16} & OSIE & 0.855 & 0.479 & 1.859 & 0.863 & 0.650 & 0.560 & 1.546 & 0.886 \\
      & Salicon $\to$ OSIE & 0.875 & 0.553 & 1.555 & 0.884 & 0.701 & 0.668 & 1.997 & 0.686 \\
      & ImageNet $\to$ OSIE & 0.876 & 0.575 & 1.434 & 0.882 & 0.723 & 0.707 & 2.204 & 0.648\\
      & ImageNet $\to$ Salicon $\to$ OSIE & 0.891 & 0.620 & 1.268 & 0.901 & 0.733 & 0.759 & 2.399 & 0.538\\
      \hline
      \multirow{4}{*}{DenseNet-161} & OSIE & 0.771 & 0.381 & 2.479 & 0.784 & 0.594 & 0.383 & 1.011 & 1.225\\
      & Salicon $\to$ OSIE & 0.849 & 0.512 & 1.756 & 0.859 & 0.662 & 0.604 & 1.772 & 0.814 \\
      & ImageNet $\to$ OSIE & 0.879 & 0.581 & 1.443 & 0.887 & 0.727 & 0.723 & 2.248 & 0.632\\
      & ImageNet $\to$ Salicon $\to$ OSIE & 0.897 & 0.647 & 1.192 & 0.904 & 0.739 & 0.800 & 2.584 & 0.483\\
      \hline
      \multirow{4}{*}{DPN-131} & OSIE & 0.830 & 0.457 & 2.019 & 0.841 & 0.621 & 0.514 & 1.410 & 0.966\\
      & Salicon $\to$ OSIE & 0.847 & 0.500 & 1.830 & 0.858 & 0.651 & 0.595 & 1.738 & 0.842 \\
      & ImageNet $\to$ OSIE & 0.886 & 0.625 & 1.318 & 0.893 & 0.731 & 0.768 & 2.482 & 0.565\\
      & ImageNet $\to$ Salicon $\to$ OSIE & 0.898 & 0.654 & 1.162 & 0.904 & 0.737 & 0.799 & 2.591 & 0.483\\
      \hline
    \end{tabular}
  }
  \newline \newline
  \subcaption{All the metrics for Figure \ref{fig:upsamplings}: Comparison of upsampling-layer types in Readout Net on PASCAL-S}
  \label{tb:upsamplings-all}
  \scalebox{0.99}{
    \begin{tabular}{| c | l | l |c c c c c c c c|} \hline
      Model & \multicolumn{2}{|c|}{Readout Net} & AUC-Judd $\uparrow$ & SIM $\uparrow$ & EMD $\downarrow$ & AUC-Borji $\uparrow$ & sAUC $\uparrow$ & CC $\uparrow$ & NSS $\uparrow$ & KL $\downarrow$\\ \hline \hline
      \multirow{4}{*}{VGG-16} & \multicolumn{2}{|c|}{N=0} & 0.891 & 0.620 & 1.268 & 0.901 & 0.733 & 0.759 & 2.399 & 0.538\\
      \hhline{|~|-|-|--------|}
      & \multirow{3}{*}{Bilinear} & N=1 & 0.887 & 0.607 & 1.320 & 0.895 & 0.733 & 0.744 & 2.334 & 0.559\\
      & & N=2 & 0.883 & 0.589 & 1.375 & 0.891 & 0.730 & 0.721 & 2.248 & 0.599\\
      & & N=3 & 0.886 & 0.595 & 1.368 & 0.894 & 0.735 & 0.735 & 2.282 & 0.575\\
      \hline
      \multirow{10}{*}{DenseNet-161} & \multicolumn{2}{|c|}{N=0} & 0.897 & 0.647 & 1.192 & 0.904 & 0.739 & 0.800 & 2.584 & 0.483\\
      \hhline{|~|-|-|--------|}
      & \multirow{3}{*}{Bilinear} & N=1 & 0.893 & 0.637 & 1.226 & 0.899 & 0.742 & 0.786 & 2.535 & 0.513\\
      & & N=2 & 0.895 & 0.641 & 1.200 & 0.904 & 0.742 & 0.795 & 2.562 & 0.494\\
      & & N=3 & 0.894 & 0.640 & 1.226 & 0.900 & 0.742 & 0.789 & 2.540 & 0.504\\
      \hhline{|~|-|-|--------|}
      & \multirow{3}{*}{Deconvolution} & N=1 & 0.894 & 0.651 & 1.190 & 0.901 & 0.735 & 0.792 & 2.569 & 0.498\\
      & & N=2 & 0.887 & 0.629 & 1.313 & 0.895 & 0.740 & 0.770 & 2.513 & 0.545\\
      & & N=3 & 0.892 & 0.635 & 1.254 & 0.899 & 0.733 & 0.776 & 2.493 & 0.520\\
      \hhline{|~|-|-|--------|}
      & \multirow{3}{*}{Subpixel Convolution} & N=1 & 0.894 & 0.639 & 1.216 & 0.901 & 0.739 & 0.788 & 2.561 & 0.504\\
      & & N=2 & 0.891 & 0.632 & 1.252 & 0.899 & 0.742 & 0.777 & 2.519 & 0.525\\
      & & N=3 & 0.895 & 0.645 & 1.192 & 0.902 & 0.741 & 0.793 & 2.557 & 0.495\\
      \hline
      \multirow{10}{*}{DPN-131} & \multicolumn{2}{|c|}{N=0} & 0.898 & 0.654 & 1.162 & 0.904 & 0.737 & 0.799 & 2.591 & 0.483\\
      \hhline{|~|-|-|--------|}
      & \multirow{3}{*}{Bilinear} & N=1 & 0.896 & 0.631 & 1.236 & 0.903 & 0.729 & 0.780 & 2.501 & 0.506\\
      & & N=2 & 0.893 & 0.630 & 1.261 & 0.901 & 0.727 & 0.779 & 2.490 & 0.526\\
      & & N=3 & 0.895 & 0.632 & 1.242 & 0.903 & 0.735 & 0.787 & 2.533 & 0.509\\
      \hhline{|~|-|-|--------|}
      & \multirow{3}{*}{Deconvolution} & N=1 & 0.900 & 0.653 & 1.188 & 0.907 & 0.739 & 0.793 & 2.562 & 0.478\\
      & & N=2 & 0.894 & 0.642 & 1.245 & 0.900 & 0.737 & 0.782 & 2.522 & 0.518\\
      & & N=3 & 0.895 & 0.645 & 1.216 & 0.902 & 0.735 & 0.785 & 2.523 & 0.500\\
      \hhline{|~|-|-|--------|}
      & \multirow{3}{*}{Subpixel Convolution} & N=1 & 0.897 & 0.658 & 1.150 & 0.905 & 0.742 & 0.799 & 2.585 & 0.496\\
      & & N=2 & 0.897 & 0.644 & 1.214 & 0.904 & 0.739 & 0.788 & 2.549 & 0.491\\
      & & N=3 & 0.890 & 0.629 & 1.317 & 0.898 & 0.733 & 0.769 & 2.480 & 0.533\\
      \hline
    \end{tabular}
  }
\end{table*}

\end{document}